\documentclass{article}

\usepackage{arxiv}

\usepackage[utf8]{inputenc} 

\usepackage{amsfonts}       
\usepackage{nicefrac}       
\usepackage{microtype}      
\usepackage{lipsum}

\usepackage{color}
\usepackage{listings}
\usepackage{booktabs}
\usepackage{array}
\usepackage{graphicx}
\usepackage{longtable}
\usepackage{subfigure}

\usepackage{cite}
\usepackage{url}
\usepackage{fancyhdr}

\usepackage{soul}

\usepackage{float}
\usepackage{listings}
\usepackage{mdwmath}
\usepackage{mdwtab}
\usepackage{multirow}
\usepackage{multicol}
\usepackage{rotating}
\usepackage{setspace}
\usepackage[utf8]{inputenc}
\usepackage{lineno}
\usepackage{listings}

\usepackage{mdwmath}
\usepackage{mdwtab}
\usepackage{multirow}
\usepackage{multicol}
\usepackage{array}
\usepackage{booktabs}

\usepackage{enumitem}
\usepackage{xspace}
\usepackage[export]{adjustbox}
\usepackage{graphicx}
\usepackage{color,soul}
\usepackage{rotating}
\usepackage{setspace}
\usepackage{amsmath} 
\usepackage{amssymb}
\usepackage{float}
\usepackage{xcolor}

\usepackage{hyperref}
\usepackage[numbers]{natbib}
\usepackage{url}

\title{Responsible-AI-by-Design: a Pattern Collection for Designing Responsible AI Systems}

\author{Qinghua Lu, Liming Zhu, Xiwei Xu, Jon Whittle\\
Data61, CSIRO, Australia}

\begin{document}

\maketitle

\begin{abstract}
Although AI has significant potential to transform society, there are serious concerns about its ability to behave and make decisions responsibly. Many ethical regulations, principles, and guidelines for responsible AI have been issued recently. However, these principles are high-level and difficult to put into practice. In the meantime much effort has been put into responsible AI from the algorithm perspective, but they are limited to a small subset of ethical principles amenable to mathematical analysis. Responsible AI issues go beyond data and algorithms and are often at the system-level crosscutting many system components and the entire software engineering lifecycle. Based on the result of a systematic literature review, this paper identifies one missing element as the system-level guidance — how to design the architecture of responsible AI systems. We present a summary of design patterns that can be embedded into the AI systems as product features to contribute to responsible-AI-by-design.

\end{abstract}


\section{Introduction}
\label{intro}

Although AI has significant potential and capacity to stimulate economic growth and improve productivity across a growing range of domains, there are serious concerns about the AI systems’ ability to behave and make decisions in a responsible manner.
According Gartner's recent report, 21\% of organizations have already deployed or plan to deploy responsible AI technologies within the next 12 months\footnote{\url{https://www.gartner.com/en/articles/it-budgets-are-growing-here-s-where-the-money-s-going}}. 

Many ethical principles, and guidelines have been recently issued by governments, research institutions, and companies~\cite{jobin2019global}. 
However, these principles are high-level and can hardly be used in practice by developers. Responsible AI research has been focusing on algorithm solutions limited to a subset of issues such as fairness\cite{mehrabi2021survey}.  
Ethical issues can enter at any point of the software engineering lifecycle and are often at the system-level crosscutting many components of AI systems. 
To try to fill the principle-algorithmic gap, some development guidelines have started to appear. However, those efforts tend to be high-level development process checklists\footnote{\url{https://ec.europa.eu/info/funding-tenders/opportunities/docs/2021-2027/horizon/guidance/ethics-by-design-and-ethics-of-use-approaches-for-artificial-intelligence_he_en.pdf}}\footnote{\url{https://standards.ieee.org/wp-content/uploads/import/documents/other/ead_v2.pdf}} and ad-hoc sets lacking of state-related linkages for final products~\cite{lu2022software}.


Therefore, in this paper, rather than staying at the ethical principle-level or AI algorithm-level, we take a pattern-oriented approach and focuses on the system-level design patterns to build responsible-AI-by-design into final AI products. The design patterns are collected based on the results of a systematic literature review (SLR) and can be embedded into the design of AI systems as product features to contribute to responsible-AI-by-design. We identify the lifecycle of a provisioned AI system in which the states or state transitions are associated with design patterns to show when the design patterns can take effects. The lifecycle along with the design pattern annotations provide an responsible-AI-focused view of system interactions and a guide to effect use of design patterns to implement responsible AI from a system perspective. To the best of our knowledge, this is the first study that provides a concrete and actionable system-level design guidance for architects and developers to reference. 


\begin{figure*}
\centering
\includegraphics[width=0.9\textwidth]{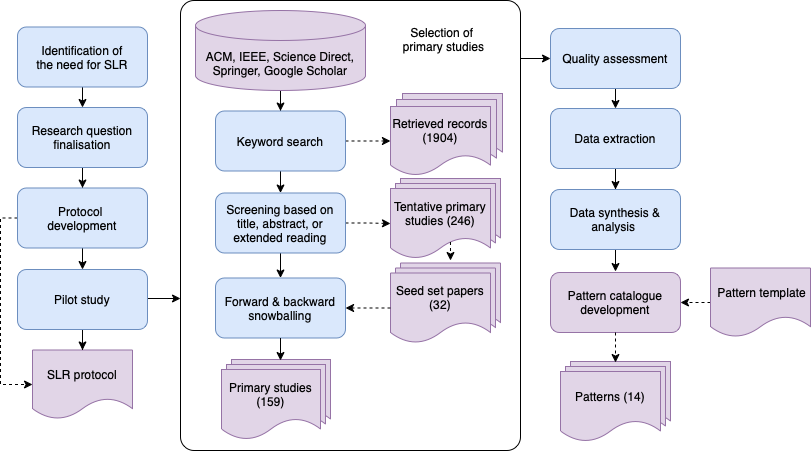}
\caption{Methodology.} \label{fig:methodology}
\vspace{-2ex}
\end{figure*}


\section{Methodology}
To operationalize responsible AI, we performed an SLR to identify design patterns that architects and developers can use during the development process.
Fig.~\ref{fig:methodology} illustrates the methodology. The research question is: "What solutions for responsible AI can be identified?" The research question focuses on identifying the reusable patterns for responsible AI.
We used "AI", "Responsible", "Solution" as the key terms and included synonyms and abbreviations as supplementary terms to increase the search results. 
The main data sources are ACM Digital Library, IEEE Xplore, Science Direct, Springer Link, and Google Scholar. 
The study only includes the papers that present concrete design or process solutions for responsible AI, and excludes the papers that only discuss high-level frameworks. 
The complete SLR protocol is available as online material~\footnote{\url{https://drive.google.com/file/d/1Ty4Cpj_GzePzxwov5jGKJZS5AvKzAy3Q/view?usp=sharing}}. We use the ethical principles listed in Harvard University's mapping study ~\cite{fjeld2020principled}: Privacy, Accountability (professional responsibility is merged into accountability due to the overlapping definitions), Safety \& Security, Transparency \& Explainability, Fairness and Non-discrimination, Human Control of Technology, Promotion of Human Values. 

\begin{figure*}
\centering
\includegraphics[width=0.9\textwidth]{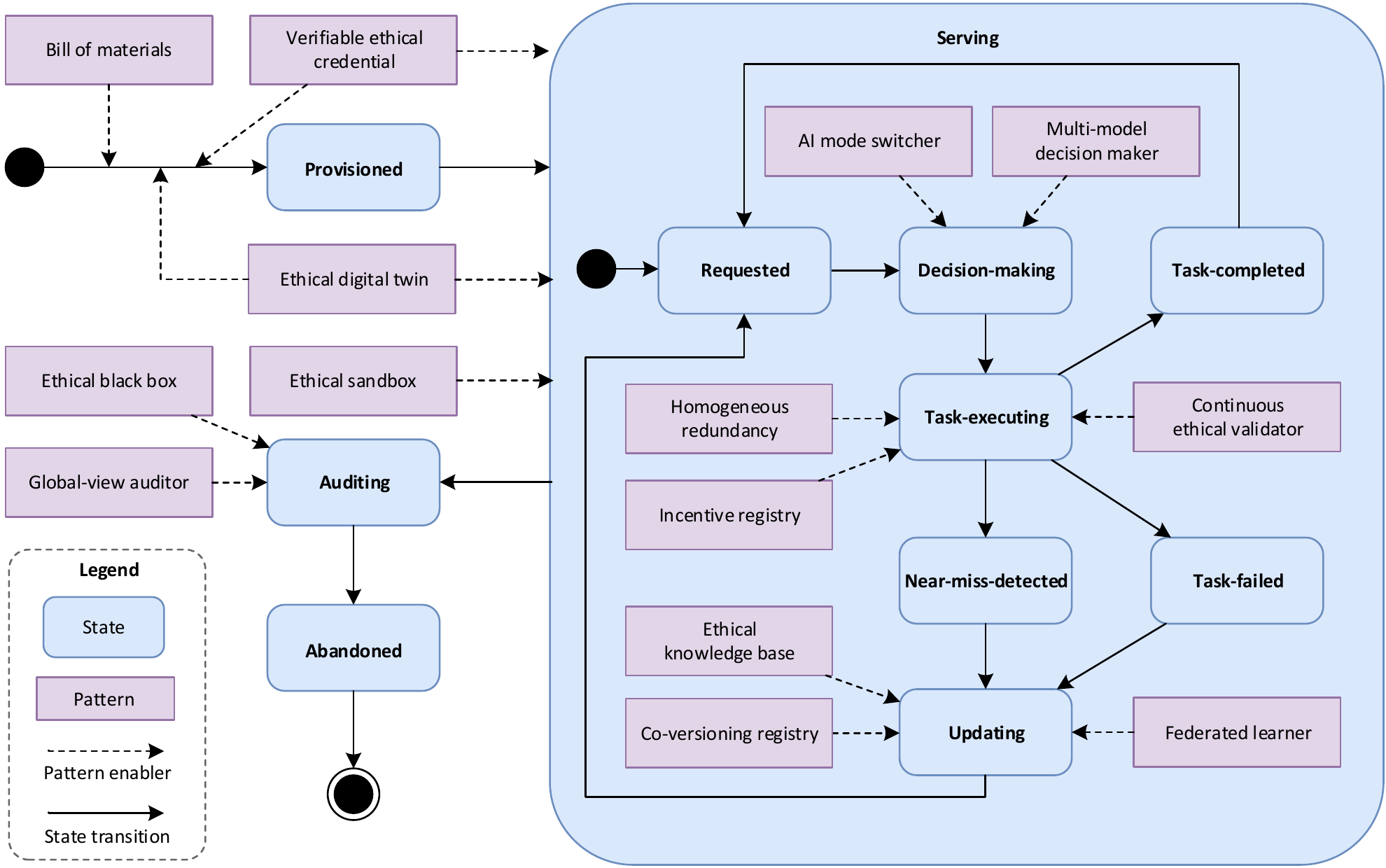}
\caption{Lifecycle of a provisioned AI system.} \label{fig:lifecycle}
\vspace{-2ex}
\end{figure*}









\section{Lifecycle of a provisioned AI System}
Fig.~\ref{fig:lifecycle} illustrates the lifecycle of a provisioned AI system using a state diagram and highlights the patterns associating with relevant states or transitions, which show when the design patterns could take effect. We have limited the scope to the design patterns that can be embedded into the AI systems and provisioned supply chain tool-chain as final product features. The best practices of the development process including some patterns related to offline model training is out of the scope of this paper. Before an AI system is provisioned, the supply chain information can be accessed through \textit{bill of materials}. The users can be required to provide the \textit{verifiable ethical credentials} to show their capability to operate the systems, while the users can examine the system's \textit{verifiable ethical credentials} for ethical compliance checking. Once the AI system starts serving, it is important to perform system-level simulation through an \textit{ethical digital twin}. \textit{ethical sandbox} can be used to physically separate AI components from non-AI components. When an AI system is requested to execute a task, \textit{decision-making} is often needed before executing the task. AI component can be activated or deactivated through \textit{AI model switcher} to automatically make the decision or involve human experts to review the suggestion. \textit{Multi-model decision maker} can use different models to make a single decision and cross-check the results. Similarly, \textit{homogeneous redundancy} can be applied to the system design to enable fault-tolerance. Both the behaviors and decision-making outcomes of the AI system are monitored and validated through \textit{continuous ethical validator}. Incentives for the ethical behaviors can be maintained by an \textit{incentive registry}. If the system is failed to meet the requirements (including ethical requirements) or a near-miss is detected, the system need to be updated. \textit{Federated learner} retrains the model locally at each client to protect data privacy. \textit{Co-versioning registry} can be used to track the co-evolution of AI system components or assets. An \textit{ethical knowledge base} can be built to make the ethical knowledge systematically accessed and used when developing or updating the AI system. The AI system needs to be audited regularly or when major-failures/near-misses occur. An \textit{ethical black box} can be designed to record the critical data that can be kept as evidence. A \textit{global-view auditor} can be built on top to provide global-view accountability when multiple systems are involved in an accident. The stakeholders can determine to abandon the AI system if it no longer fulfils the requirements.



\begin{figure*}
\centering
\includegraphics[angle=90, width=0.95\textwidth]{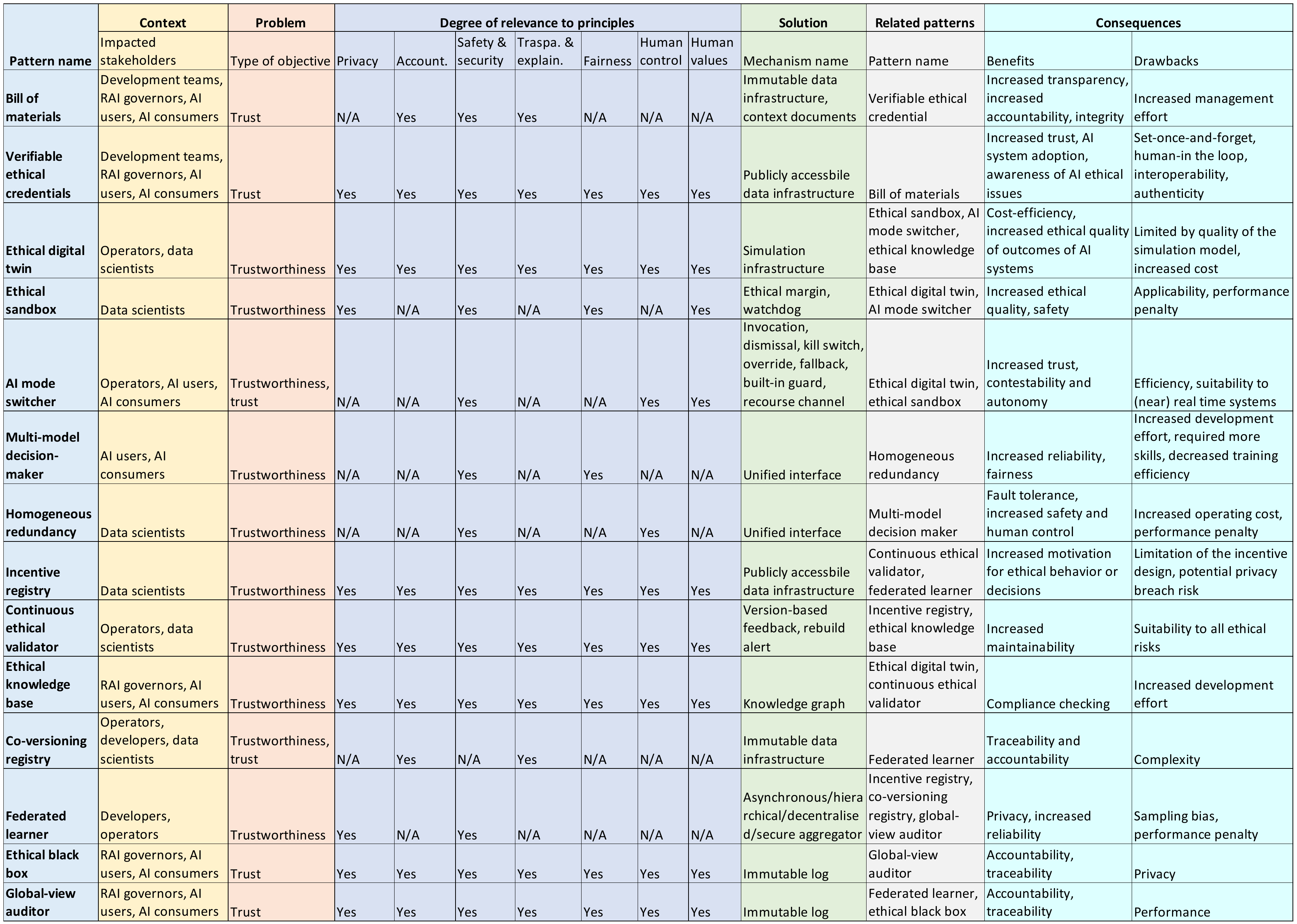}
\caption{Operationalized design patterns for responsible AI systems.}
\label{fig:patterns}
\vspace{-2ex}
\end{figure*}

\section{Design Patterns}
To operationalize responsible AI, 
Fig.~\ref{fig:patterns} lists a collection of patterns for responsible-AI-by-design. The full version of design patterns is available online~\footnote{ \url{https://drive.google.com/file/d/1SBuqkdx91hzcxiGjtxxMyt_55JzlVBK6/view?usp=sharing}}. 

\begin{itemize}
    \item \textbf{Bill of materials}: AI product vendors often create AI systems by assembling commercial or open source AI and/or non-AI components from third parties. The AI users often have ethical concerns about the procured AI systems/components. Before an AI system is provisioned, the supply chain information can be accessed through \textit{bill of materials}~\footnote{ \url{https://www.ntia.doc.gov/files/ntia/publications/sbom_minimum_elements_report.pdf}}, which  keeps a formal machine-readable record of the supply chain details of the components used in building an AI system, 
    such as component name, version, supplier, dependency relationship, author and timestamp. The purpose of bill of material is to provide traceability and transparency into the components that make up AI systems so that ethical issues can be tracked and addressed. There have been many tools to generate SBOM for practitioners, such as Dependency-Track~\footnote{\url{https://dependencytrack.org}}. To ensure traceability and integrity, \textit{immutable data infrastructure}~\cite{barclay2022providing} is needed to store the data of bill of materials. For example, the manufacturers of autonomous vehicles can maintain a material registry contract on blockchain to track their components' supply chain information, e.g., the version and supplier of the third-party AI-based navigation component. 
    
    \item \textbf{Verifiable ethical credential}: \textit{Verifiable ethical credentials} are cryptographically verifiable data that can be used as strong proof of ethical compliance for AI systems, components, artifacts, and stakeholders (such as developers and users). Before using the provisioned AI systems, users verify the systems' ethical credential to check if the systems are compliant with AI ethics principles or regulations~\cite{chu2021}. On the other hand, the users is often required to provide the ethical credentials to use and operate the AI systems. 
    \textit{Publicly accessible data infrastructure} needs to be built to support the generation and verification for ethical credentials~\footnote{\url{https://securekey.com}}. 
    For example, before driving an vehicle, the driver is requested to scan her/his ethical credential to show she/he has the capability to drive safely, while verifying the ethical credential of the vehicle's automated driving system shown on the center console. 
    
    \item \textbf{Ethical digital twin}: Before running the provisioned AI system in a production environment, it is critical to conduct system-level simulation through an \textit{ethical digital twin} running on a simulation platform to monitor the behaviors of AI systems and predict potential ethical risks. Ethical digital twin can also be designed as a component at the operation infrastructure level to examine the AI systems' runtime behaviors and decisions based on the abstract simulation model using the real-world data. The risk assessment results can be used by the system or users to take further actions to mitigate the potential ethical risk. 
    For example, the manufacturers of autonomous vehicles can use the ethical digital twin to explore the limits of autonomous vehicles based on the collected run-time data, such as NVIDIA DRIVE Sim~\footnote{\url{https://developer.nvidia.com/drive/drive-sim}} and xFpro~\footnote{\url{https://rfpro.com}}. 
    
    \item \textbf{Ethical sandbox}:
    It is risky to execute the whole system including AI components and non-AI components in the same environment. When an AI system is being served, \textit{ethical sandbox} can be used to physically separate AI components from non-AI components by running the AI component in a self-contained emulation execution environment~\cite{lavaei2021towards}, e.g. sandboxing the unverified visual perception component. 
    The AI components placed in the ethical sandbox has no access to the rest of the AI system. All the hardware and software functionality of the AI component are duplicated in the ethical sandbox. Thus, the AI component can run safely under supervision before being deployed at scale. For example, Fastcase AI Sandbox~\footnote{\url{https://www.fastcase.com/sandbox/}} provides a secure AI execution platform for analysing data safely in a secure environment.
    Maximal tolerable probability should be set as an \textit{ethical margin} for the sandbox against the ethical requirements. 
    A \textit{watch dog} can be added to restrict the execution time of the AI component to avoid the potential ethical risk, e.g., only executing the visual perception component for 10 minutes on the roads designed especially for autonomous vehicles. 
    
    \item \textbf{AI mode switcher}: When to activate AI is a major architectural design decision when designing a software system. When an AI system is making a decision, \textit{AI mode switcher} enables efficient \textit{invocation and dismissal} mechanisms for activating or stopping the AI component when needed. \textit{Kill switch} is a special type of invocation mechanism which immediately turns off the AI component and terminates its negative effects, e.g. switching off the autopilot functionality~\footnote{ \url{https://www.tesla.com/autopilot}} and its internet connection. The AI component can make decisions automatically or provide suggestions to human experts in high risk situations. The decisions can be approved or overridden by human expert (e.g. skipping the path suggested by the navigation system). If the system state after acting an AI decision is not expected by human experts, \textit{fallback} can be triggered to reverse the system back to the previous state. A \textit{built-in guard} ensures that the AI component is only being used under the predefined risk categories.  
    
    \item \textbf{Multi-model decision-maker}: The reliability of traditional software is dependent on the design of software components. One of the reliability practices in the reliability community is redundancy, which can be applied to AI components. When decisions are being made by an AI system, \textit{multi-model decision-maker} can run different models to make a single decision~\cite{TFUtils_Multi-Model_Training}, e.g., using different algorithms for visual perception. 
    Reliability can be improved by using different models under different context (e.g., different user groups or regions). In addition, fault tolerance can be enabled by cross-checking the results given by multiple models (e.g., only accepts the same results from the deployed models). IBM Watson Natural Language Understanding make predictions using an ensemble learning framework that includes multiple emotion detection models~\footnote{ \url{https://www.ibm.com/au-en/cloud/watson-natural-language-understanding}}. 
    
    \item \textbf{Homogeneous redundancy}:  Ethical failures in AI systems can cause serious damage to the humans or environment. N-version programming is a design pattern for dealing with reliability issues of traditional software. This concept can be adapted and applied to AI system design. 
    \textit{Homogeneous redundancy} (e.g., two brake control components) can be applied to tolerate the highly uncertain AI system components that can make unethical decisions or the adversary hardware components that produce malicious data or behave unethically~\cite{tidjon2022threat}. When an AI system is executing a task, a cross-check can be performed for the outputs given by multiple redundant components of a single type. 

    \item \textbf{Incentive registry}:
    Incentives are effective in motivating AI systems to execute tasks in a responsible manner.
    When executing a task, an \textit{incentive registry} records the rewards that are given for the behavior and decisions and behaviors of AI systems~\cite{deepchain}, e.g., rewards for the recommended path without safety risk.
    There are different ways to enforce the incentive mechanism, e.g., designing the incentive mechanism on blockchain based \textit{data infrastructure that is publicly accessible}, using reinforcement learning.  	  
    However, it is challenging to design the mechanisms in the responsible AI context since it difficult to measure the ethical impact of decisions and behaviors of AI systems on some ethical principles (such as human values). Besides, consensus needs to be reached on the incentive mechanism by all the stakeholders. Additionally, in some cases, ethical principles are conflicting with each other, making the design of incentive mechanism harder. FLoBC~\footnote{\url{https://github.com/Oschart/FLoBC}} is a tool that uses blockchain to incentivize training contribution for federated learning.

    \item \textbf{Continuous ethical validator}: 
    AI systems often need to conduct continual learning when data drift or unethical behavior is detected in production. When an AI system executes tasks, 
    \textit{continuous ethical validator} monitors and validates the outcomes of AI systems (e.g., the path suggested by the navigation system) against the ethical requirements. 
    The outcomes of AI systems are the consequences of decisions and behaviors of the systems, i.e., whether the AI system behaves ethically or provides the promised benefits in a given situation.
    The time and frequency of validation can be predefined within the continuous validator. \textit{Version-based feedback} and \textit{rebuild alert} can be sent when the ethical requirements are met or breached. \textit{Incentive registry} can be used to reward or punish the ethical/unethical behavior or decisions of AI systems. 
    
    \item \textbf{Ethical knowledge base}:
    AI systems involve broad ethical knowledge, including AI ethics principles, regulations, unethical use cases, etc. Unfortunately, such ethical knowledge is scattered in different documents (e.g., AI incidents) and is usually implicit or even unknown to developers who primarily focus on the technical aspects of AI systems and do not have ethics background. This results in negligence or ad-hoc use of relevant ethical knowledge in AI system development. Ethical knowledge base is built upon a \textit{knowledge graph} to make meaningful entities, concepts and their rich semantic relationships are explicit and traceable across heterogeneous documents so that the ethical knowledge can be systematically accessed, analysed, used when developing or updating AI systems~\cite{Naja21}. For example, an ethical knowledge base can be used to support continuous ethical risk assessment. Ethical knowledge base can be built based on the AI ethics principles, frameworks, and actual AI use cases discussed in the existing papers. 

    \item \textbf{Co-versioning registry}:
    AI systems involve different levels of dependencies and need frequent evolution when data drift or unethical behavior occurs. Co-versioning of the components of AI systems or AI assets generated in AI pipelines provides provenance guarantees across the entire lifecycle of AI systems. There have been many version control tools for managing the co-versioning of data and models, such as DVC~\footnote{ \url{https://dvc.org/}}. When updating an AI system, \textit{co-versioning registry} can track the co-evolution of components or AI assets. There are different levels of co-versioning: co-versioning of AI components and non-AI components, co-versioning of the assets within the AI components (i.e., co-versioning of data, model, code, configurations). A publicly accessible data infrastructure can be used to maintain the co-versioning registry to provide a trustworthy trace for dependencies. For example, a co-versioning registry contract can be built on blockchain to manage different versions of visual perception models and the corresponding training datasets.

    \item \textbf{Federated learner:} 
    Despite the widely deployed mobile or IoT devices generating massive amounts of data, data hungriness is still a challenge given the increasing concern in data privacy. When learning or updating AI models, \textit{federated learner} preserves the data privacy by performing the model training locally on the client devices and formulating a global model on a central server based on the local model updates~\footnote{\url{https://leaf.cmu.edu}}, e.g., train the visual perception model locally in each vehicle. 
    \textit{Decentralized learning} is an alternative to federated learning, which uses blockchain to remove the single point of failure and coordinate the learning process in a fully decentralized way.
    In the event of negative outcomes, the responsible humans can be traced and 
    identified by an \textit{ethical black box} for accountability. 

    \item \textbf{Ethical black box}: 
    Black box was introduced initially for aircraft several decades ago for recording critical flight data. The purpose of embedding an ethical black box in an AI system is to audit an AI system and investigate why and how the system caused an accident or a near miss. 
    The ethical black box continuously records  sensor data, internal status data, decisions, behaviors (both system and operator) and effects~\cite{falco2020distributedblack}. For example, an ethical black box could be built into the automated driving system to record the behaviors of the system and driver and their effects. 
    Design decisions need to be made on what data should be recorded and where the data should be stored (e.g. using a blockchain-based immutable log or a cloud-based data storage). 
    
    \item \textbf{Global-view auditor}:
    There can be more than one AI systems involved in an ethical incident (e.g. multiple autonomous vehicles in a car accident). During auditing, it is often challenging to identify the liability as the data collected from each of the involved systems can be conflicting to each other. 
    \textit{Global-view auditor} can enable accountability by analysing the data discrepancies between the involved AI systems and identifying the liability for the ethical incident. This pattern can be also applied to improve the reliability an AI system by taking the data from other systems.
    For example, an autonomous vehicle increases its visibility using the perception data collected from the other vehicles.~\cite{miguel2021putting}. All the historical data of AI systems can be recorded by an \textbf{immutable log} for third-party auditing.

\end{itemize}

\section{Conclusion}
To operationalize responsible AI, we take a pattern-oriented approach and collect a set of product design patterns that can be embedded into an AI system as product features to enable responsible-AI-by-design. The patterns are associated to the states or state transitions of a provisioned AI system, serving as an effective guidance for architects and developers to design a responsible AI system. We are currently building up a responsible AI pattern catalogue that includes multi-level governance patterns, trustworthy process patterns (i.e., best practices and techniques), and responsible-AI-by-design product patterns.




\end{document}